\documentclass[conference,letterpaper]{IEEEtran}

\usepackage{graphicx}
\usepackage{amsmath}
\usepackage{booktabs}
\usepackage{xcolor}
\usepackage{soul}

\usepackage[hidelinks]{hyperref}

\begin{document}

\title{Target Selection with a Disappearing-Stimulus Mixed-Reality P300 Interface:
Feasibility and Object-Distinguishable Versus Uniform Cues}

\author{\IEEEauthorblockN{Runqi Pan$^{1,2}$, Takuto Niinuma$^{2}$, Yunshan Huang$^{2}$, Masato Sugino$^{2}$, Kenta Shimba$^{2}$, and Kiyoshi Kotani$^{2,\dagger}$}
\IEEEauthorblockA{$^{1}$Imperial College London, London, UK\\
$^{2}$Graduate School of Frontier Sciences, The University of Tokyo, Kashiwa, Japan\\
$^{\dagger}$Corresponding author: Kiyoshi Kotani}}

\maketitle

\begin{abstract}
Mixed-reality (MR) P300 interfaces can register selectable destinations, such as
wheelchair goals, directly in the user's surroundings. When the cues disappear
between flashes, however, the user must retain an intended target that is no
longer visible. We tested whether such an interface supports target selection
and whether giving each cue a stable colour--shape identity improves it compared
with visually uniform spheres. Six healthy adults completed four trials in each
cue condition of a $5\times5$ row/column task on a head-mounted MR display, with
EEG from Fz, Cz, and Pz. Shrinkage linear discriminant analysis was evaluated
with nested validation in which complete trials were held out from class
balancing, hyperparameter selection, and fitting. Both conditions evoked a
parietal target-related positivity. The decoded cell was correct in 18 of 48
held-out trials, against 1.9 expected by chance (binomial $p<10^{-12}$), and in
three participants individually. Cue format made no detectable difference:
complete-target accuracy was 9/24 in both conditions (paired difference 0.0
percentage points, 95\% CI $[-37, 37]$), and held-out AUC was $0.609$ for spheres
and $0.585$ for object cues. Disappearing-stimulus MR P300 selection is therefore
feasible under strict trial-level evaluation, but the benefit of object identity,
if any, is too small to resolve in a pilot of this size.
\end{abstract}

\section{Introduction}

Brain--computer interfaces (BCIs) decode information from brain responses to
visual \cite{farwell1988,sugino2025reservoir} and auditory
\cite{senda2024auditory,tanaka2025cfc} stimuli and from motor imagery
\cite{huang2025object}. P300 is an event-related potential (ERP) elicited by
task-relevant stimuli and has been widely used as a control signal in BCIs \cite{polich2007,wolpaw2002}. In assistive wheelchairs, P300 BCIs are
often used for high-level destination selection while an autonomous layer
handles navigation \cite{iturrate2009,rebsamen2010}. Mixed reality (MR) extends
this idea by placing the selectable destinations in the user's own environment
instead of on a separate monitor. Head-mounted P300 communication
\cite{kathner2015rapid} and broader BCI--AR integration
\cite{simohammed2020arbci} have been demonstrated, immersive presentation can
change ERP amplitude and classification performance \cite{arif2026immersive},
and online MR destination selection has been used to drive a wheelchair
\cite{mori2024}.

Showing every destination continuously can clutter the scene. An alternative
hides all cues after a target preview and reveals them only as transient row
and column flashes. The user must then retain the intended destination and
recognise its brief reappearance. Two questions follow. First, does this
disappearing-stimulus design still yield usable target evidence? Second, does
the cue format matter? With identical spheres, location is the only cue to the
target; with unique colour--shape objects, the target can also be recognised by
its appearance. Visual P300 performance is known to depend on display design,
including flash grouping \cite{townsend2010}, face overlays
\cite{kaufmann2011faces}, and stimulus colour or shape
\cite{ryan2017color,fernandez2019shape}, but in the matrix-based and head-mounted P300 displays cited here the selectable items remain
visible between flashes \cite{townsend2010,kathner2015rapid,kim2021drone},
including the MR wheelchair interface of \cite{mori2024}; in
rapid serial visual presentation (RSVP) designs, items are shown one at a time
without a persistent spatial layout \cite{acqualagna2013rsvp,chennu2013}. In the
present design, the spatial layout is previewed and then removed, so that each
row or column is visible only during its 200-ms flash.
Fig.~\ref{fig:paradigm}c contrasts this design with a conventional P300
display.

Small EEG datasets add a methodological risk. Neighbouring epochs from one
trial are dependent, and random epoch-level partitions can inflate accuracy
\cite{morenoalcayde2024}. We therefore evaluated both cue formats with nested
validation in which complete trials were held out. The study has two aims: (1) to test whether P300 target selection
remains feasible when the cues disappear between flashes, and (2) to estimate
the effect of object-distinguishable colour--shape cues compared with uniform
spheres in this setting, together with its uncertainty.

\section{Methods}

\subsection{Participants and design}

Six healthy adults (five male; 20--28 years; right-handed; normal or corrected
vision; all with prior BCI experience) took part after giving written informed
consent. Two participants are authors (P1 and P6). P1 was aware of the
hypothesis; results are also reported without P1 and without both. P6 is also
a co-author of the MR wheelchair system study \cite{mori2024}. This study was conducted in accordance with the Regulations on Research
Ethics of the University of Tokyo and was approved by the Ethics Committee of
the University of Tokyo. Each participant completed four
trials with uniform spheres and four with object-distinguishable cues, giving
48 trials. Condition order was recorded but not formally counterbalanced: by file time
stamps, three participants did all sphere trials first, one did all object
trials first, P5 recorded S, O, S blocks of 2, 4 and 2 trials, and P6 alternated.

\subsection{MR task and cue conditions}

\begin{figure*}[t]
\centering
\includegraphics[width=0.86\textwidth]{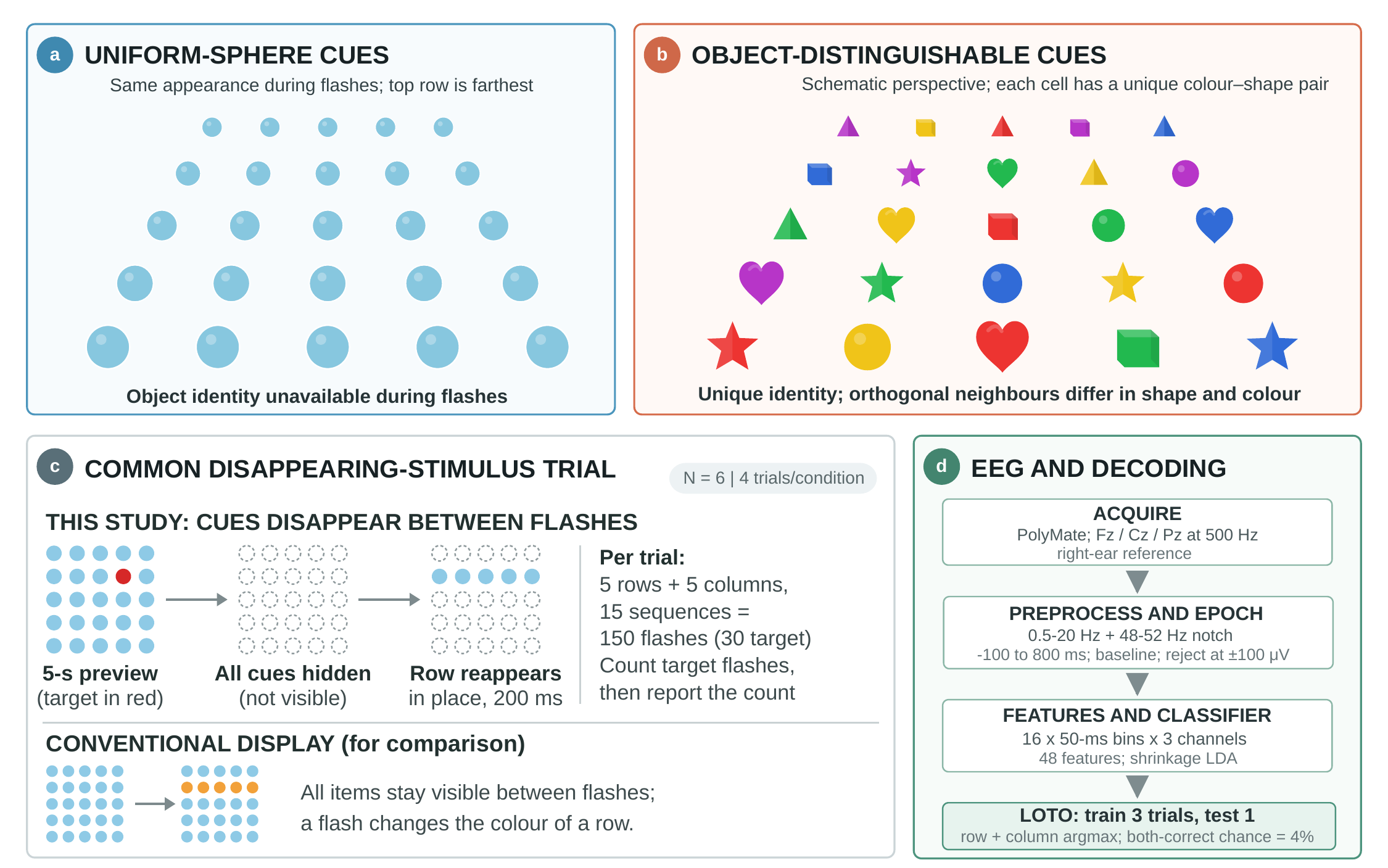}
\caption{Cue conditions, trial structure, and analysis. (a) Uniform spheres.
(b) Object-distinguishable cues; each cell has a unique colour--shape identity.
(c) Top: after a 5-s target preview all cues are hidden and reappear in place
only as 200-ms row and column flashes. Bottom: in a conventional display all
items stay visible and a flash changes the colour of a row (orange). (d) Identical acquisition, preprocessing, features, and
complete-trial-held-out decoding in both conditions.}
\label{fig:paradigm}
\end{figure*}

An HTC Vive XR Elite displayed a $5\times5$ grid of virtual destinations in the
room, with the nearest row about 2~m away (column and row spacing 0.60 and
0.45~m). In the sphere condition all cells were identical spheres. In the
object condition each cell had a unique combination of five colours and five
shapes, and orthogonal neighbours differed in both (Fig.~\ref{fig:paradigm}).
The experimenter entered the target cell for each trial; targets were not
randomised or matched between conditions. The target was previewed for 5~s,
then all cues disappeared. Rows and columns reappeared for 200~ms in
pseudo-random order, each once per sequence, for 15 sequences (150 events, 30
targets per trial; median marker interval 805.6~ms). Participants silently
counted target flashes.

\subsection{EEG recording and preprocessing}

EEG was recorded from Fz, Cz, and Pz at 500~Hz (PolyMate Mini AP108, Miyuki
Giken), referenced to the right earlobe. Data were filtered with zero-phase
Butterworth filters (0.5~Hz high-pass, 48--52~Hz band-stop, 20~Hz low-pass) and
epoched from $-100$ to 800~ms around software markers, with baseline correction
over $-100$ to 0~ms. Epochs with any sample at or above $100~\mu$V in absolute
value were rejected. The P300 was quantified as the 250--600~ms mean of the Pz
target-minus-non-target difference.

\subsection{Decoding and evaluation}

Features were the mean amplitudes of sixteen 50-ms bins from 0 to 800~ms on
each channel (48 features) \cite{krusienski2008}, classified with shrinkage
linear discriminant analysis \cite{blankertz2011}. For each participant and
condition, the outer loop held out one complete trial. Within the remaining
three trials, the shrinkage parameter (0--1) was chosen by an inner
leave-one-trial-out loop. Each fit undersampled non-targets to a 2:1 ratio with
a fixed seed. No epoch of a test trial was used for balancing, model selection,
or fitting. Row and column scores were the mean target-class score over their
accepted flashes. The decoded cell was the highest-scoring row and column, and
it was also recomputed after each of the 15 sequences. A row or column with no
accepted epoch was excluded from the choice. As a secondary analysis,
the same nested procedure was repeated with both conditions pooled, so that
each model was trained on the other seven trials of a participant.

\subsection{Statistics}

Feasibility was tested at the trial level: under the null hypothesis of no
target information, each held-out trial is correct with probability 1/25, so
the number of correct trials follows a binomial distribution. Because target
cells were not uniformly distributed, we also report a conservative test with
chance equal to the product of the most frequent target row and column
proportions, taking the larger of the two conditions
(spheres $9/24\times7/24=0.109$; object 0.097). We also ran a permutation test that preserves the target distribution:
target cells were shuffled among each participant's trials (200
permutations). Cue-format contrasts used the participant as the unit,
with two-sided exact paired sign-flip tests over all $2^6$ sign assignments
and t-based 95\% confidence intervals. With six participants the smallest
attainable exact $p$ is 0.031, so non-significant contrasts are not read as
equivalence. Complete-target accuracy was the focal outcome; Pz amplitude and
AUC were supporting outcomes. Analyses used MATLAB.

\section{Results}

\subsection{Signal quality and ERPs}

Of 7,200 epochs, 6,614 (91.9\%) passed artefact rejection (sphere $7.6\pm8.6\%$
rejected, object $8.7\pm7.1\%$). Both conditions showed a parietal
target-related positivity peaking near 400~ms (Fig.~\ref{fig:results}a). The
Pz difference was $1.44\pm1.63~\mu$V for spheres and $1.63\pm1.33~\mu$V for
object cues. It was positive in five of six participants in each condition
(exact sign-flip test of the mean, $p=0.063$; participant-level percentile bootstrap with 10,000 resamples, 95\% CIs $[0.44, 2.70]$ and
$[0.64, 2.57]~\mu$V).

\subsection{Feasibility of target selection}

\begin{figure*}[t]
\centering
\includegraphics[width=\textwidth]{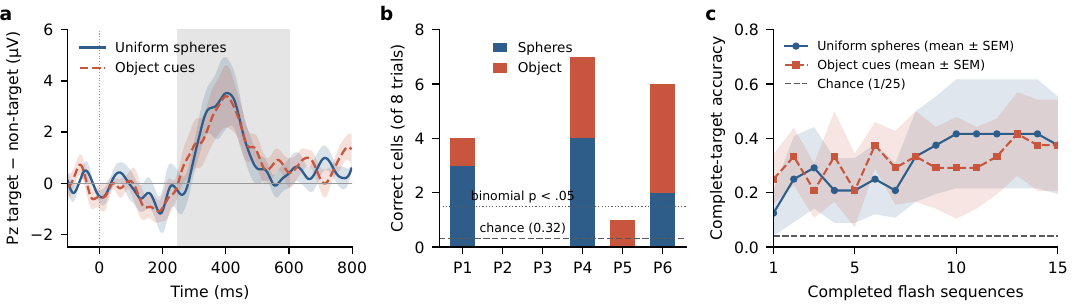}
\caption{(a) Grand-average Pz target-minus-non-target difference (mean $\pm$
SEM, $N=6$); grey marks the 250--600~ms window. (b) Correct cells out of eight
held-out trials per participant, split by condition. The dashed line is the
chance expectation (0.32); two or more correct is significant for one
participant (binomial $p<0.05$). (c) Complete-target accuracy after each
completed flash sequence (mean $\pm$ SEM across participants), decoded offline by the trial-held-out models; the dashed line is chance (0.04).}
\label{fig:results}
\end{figure*}

After 15 sequences, both the row and column were decoded correctly in 18 of 48
held-out trials (37.5\%), against 1.9 expected by chance (binomial
$p=1.6\times10^{-13}$; conservative chance 0.109, $p=1.4\times10^{-6}$).
Each condition alone reached 9/24 ($p=2.0\times10^{-7}$). Performance was
heterogeneous (Fig.~\ref{fig:results}b): three participants exceeded chance
individually (4/8, 7/8, and 6/8 correct; each $p<0.001$), one was correct once,
and two never were. Excluding the author-participant aware of the
hypothesis (P1) left 14/40 correct ($p=2.3\times10^{-10}$); excluding both
author-participants (P1 and P6) left 8/32 correct ($p=2.9\times10^{-5}$;
$p=0.062$ against the conservative chance level of these four
participants, 0.137). Row and column accuracies were 52\% and 54\% (chance
20\%). At 15 sequences one selection took about 2~min, corresponding to an
information transfer rate of 0.82~bit per selection or about 0.4~bit/min.
Training a single classifier on both cue conditions (seven trials per
model) gave 20/48 correct (spheres 11/24, object 9/24; $p=6\times10^{-16}$)
and AUC $0.605\pm0.125$. In a permutation test that
shuffled target cells among each participant's eight trials and repeated the
full nested pipeline, no permutation reached the observed 18 correct (null
mean 2.3, maximum 8; $p=0.005$, the smallest attainable with 200
permutations).

\subsection{Cue-format comparison}

Cue format produced no detectable difference in any principal outcome
(Table~\ref{tab:outcomes}). Complete-target accuracy was identical (paired
difference 0.0 percentage points, 95\% CI $[-37.1, 37.1]$). Object cues had a
descriptively higher row accuracy but a lower column accuracy, which cancelled
in the conjunction. Held-out AUC was slightly lower for object cues
(difference $-0.023$, $[-0.102, 0.055]$). Classifiers transferred across
conditions with similar performance: training on spheres and testing on object cues gave
AUC $0.593\pm0.106$, and the reverse gave $0.618\pm0.114$.

\begin{table}[t]
\caption{Participant-level outcomes ($N=6$, mean $\pm$ SD)}
\label{tab:outcomes}
\centering
\setlength{\tabcolsep}{3pt}
\resizebox{\columnwidth}{!}{%
\begin{tabular}{lcccc}
\toprule
Outcome & Spheres & Object & Difference [95\% CI] & $p$ \\
\midrule
Pz diff.\ ($\mu$V) & $1.44\pm1.63$ & $1.63\pm1.33$ & $0.19\ [-1.09, 1.46]$ & .72 \\
AUC & $.609\pm.134$ & $.585\pm.092$ & $-.023\ [-.102, .055]$ & .47 \\
Row acc.\ (\%) & $41.7\pm40.8$ & $62.5\pm30.6$ & $20.8\ [-14.0, 55.7]$ & .25 \\
Column acc.\ (\%) & $58.3\pm37.6$ & $50.0\pm38.7$ & $-8.3\ [-29.8, 13.1]$ & .63 \\
Complete target (\%) & $37.5\pm44.0$ & $37.5\pm41.1$ & $0.0\ [-37.1, 37.1]$ & 1.00 \\
\bottomrule
\end{tabular}}
\\[2pt]
\parbox{\columnwidth}{\footnotesize Differences are object minus spheres; $p$
from exact paired sign-flip tests, unadjusted and exploratory.}
\end{table}

\subsection{Accuracy across flash sequences}

Fig.~\ref{fig:results}c asks how many flash sequences a selection needs. With
hidden cues the user must hold the target in mind for the whole trial, so
evidence could keep accumulating or stop improving as attention fades; the
curve also shows the speed--accuracy trade-off for an online interface, in
which fewer sequences mean faster selections. It was computed offline: each
held-out trial was decoded by the same trial-held-out model from its first
$k$ sequences, so it does not measure online performance.
Accuracy did not increase monotonically with the number of sequences
(Fig.~\ref{fig:results}c): for spheres it was 12.5\%, 20.8\%, 41.7\%, and 37.5\%
after 1, 5, 10, and 15 sequences, and for object cues 25.0\%, 20.8\%, 29.2\%,
and 37.5\%. The curves crossed repeatedly and did not separate the cue
formats. Because accuracy for spheres was no higher at 15 than at 10 sequences,
fewer than 15 sequences may suffice, but these data do not identify an optimal
number; an online study should test this.

\subsection{Robustness}

Excluding the two trials with more than 50\% rejected epochs and retraining
left participant-mean complete-target accuracy at 37.5\% in both conditions
(17/46 trials correct; $p<10^{-11}$). Choosing any of the six
possible pairs of archived P5 sphere recordings changed the sphere AUC only
between 0.601 and 0.619. Accuracy was 11/24 in each participant's first four
trials and 7/24 in the last four. Because four participants recorded one
condition entirely before the other (of the other two, P5 recorded S, O, S blocks of 2, 4 and 2 trials and P6
alternated), this comparison is confounded with condition for those four, but
it gives no sign that practice raised accuracy.

\section{Discussion}

Hiding MR destinations between flashes still allowed P300 target selection.
Under evaluation that kept every test trial out of balancing, model selection,
and fitting, the decoded cell was correct about nine times more often than
chance, and the result held under a conservative chance level, after removing
the hypothesis-aware author-participant (P1), and after removing high-artefact
trials. With both author-participants removed, 8/32 remained above the 1/25
chance level but not above the conservative level ($p=0.062$). This matters
for MR navigation interfaces because it permits an uncluttered scene in which
destinations appear only when they are being probed.

Giving each destination a stable colour--shape identity did not yield a
detectable advantage. Object identity may have been redundant because spatial
attention already sufficed, it may have aided target retention without
enlarging the P300, or its effect may simply be smaller than six participants
can resolve; the confidence interval for complete-target accuracy spans
$\pm37$ percentage points. Colour \cite{ryan2017color} and shape \cite{fernandez2019shape} benefits reported for screen-based
spellers should therefore not be
assumed to transfer to disappearing MR cues. The similar cross-condition transfer
suggests that one classifier could serve both cue formats, although four
training trials cannot establish cue invariance.

Performance varied strongly between participants, as in large P300 cohorts
\cite{guger2009}, and the information transfer rate was low. Both follow
partly from deliberate choices for this initial test: a conservative 805-ms
marker interval suited to MR rendering, only three midline channels, and only
three calibration trials per model. Row/column paradigms also suffer when
target row and column flashes occur back to back \cite{townsend2010}. Shorter
intervals, a posterior montage, and dynamic stopping are the obvious next steps.

The study has clear limitations. The sample was small, healthy, and
BCI-experienced; condition order was not counterbalanced; target cells were
chosen by the experimenter rather than randomised; and neither eye movements
nor marker-to-photon latency \cite{swami2023clet} were measured. All outcomes
were offline. A confirmatory online study should randomise order and targets,
fix complete-target accuracy as the primary outcome, and size the sample to
detect a difference that matters for navigation.

\section{Conclusion}

A disappearing-stimulus MR P300 interface supported above-chance target
selection under strict complete-trial-held-out evaluation. Object-distinguishable
cues did not detectably outperform uniform spheres, and the wide uncertainty
around this contrast defines the sample needed for a confirmatory online study.

\section*{Acknowledgment}

The authors thank the participants and the other members of the Kotani \& Shimba Laboratory, The University of Tokyo.
A generative-AI assistant was used to support analysis-code review and
language editing; the authors verified all analyses and text and take full
responsibility for them.

\bibliographystyle{IEEEtran}
\bibliography{references}

\end{document}